\newif\ifarxiv
\arxivtrue 

\ifarxiv
\documentclass[twocolumn]{article}
\newcommand{\IEEEmembership}[1]{}
\newenvironment{IEEEkeywords}{%
  \bfseries\textit{Index Terms}---%
}{%
}

\usepackage[letterpaper, top=0.6in, bottom=0.5in, inner=0.75in, outer=0.75in, headheight=0.3in, footskip=0.3in]{geometry}
\usepackage{fancyhdr}
\usepackage{times}
\usepackage{booktabs}
\else
    \documentclass[journal]{IEEEtran}
\fi

\usepackage[pdftex]{graphicx}
\usepackage[cmex10]{amsmath}
\usepackage{algorithm}
\usepackage{algpseudocode}
\usepackage{xcolor}
\usepackage{array}
\usepackage{booktabs}
\usepackage{stfloats}
\usepackage{hyperref}
\usepackage{url}
\usepackage{verbatim}
\usepackage[sort&compress, numbers]{natbib}
\usepackage{caption}
\usepackage{tablefootnote}
\usepackage{amssymb}
\usepackage[T1]{fontenc}
\usepackage[utf8]{inputenc}
\usepackage{dirtytalk}

\begin{document}
%
\title{Chaotic Map based Compression Approach to Classification}
%
%
%

\ifarxiv
   \author{Harikrishnan~N~B\thanks{Harikrishnan~N~B is with Department of Computer Science and Information Systems, BITS Pilani K K Birla Goa Campus, 403726, Goa, India (e-mail: harikrishnannb@goa.bits-pilani.ac.in)\\
   Also at: Adjunct Faculty, Consciousness Studies Programme, National Institute of Advanced Studies, Indian Institute of Science Campus, Bengaluru, 560012.}, 
  Anuja~Vats\thanks{Anuja Vats is with the Department of Computer Science, Norwegian Institute of Science and Technology, Gjøvik, Norway (e-mail: anuja.vats@ntnu.no)}, 
  Nithin~Nagaraj\thanks{Nithin~Nagaraj is with the Complex Systems Programme, National Institute of Advanced Studies, Indian Institute of Science Campus, Bengaluru, 560012, Karnataka, India (e-mail: nithin@nias.res.in)}, 
  Marius~Pedersen\thanks{Marius Pedersen is also with the Department of Computer Science, Norwegian Institute of Science and Technology, Gjøvik, Norway (e-mail: marius.pedersen@ntnu.no)}, }
    
\else

\fi

%
%

\markboth{Journal of \LaTeX\ Class Files,~Vol.~13, No.~9, September~2014}%
{Vats \MakeLowercase{\textit{et al.}}: Bare Demo of IEEEtran.cls for Journals}
%



\maketitle

\begin{abstract}
Modern machine learning approaches often prioritize performance at the cost of increased complexity, computational demands, and reduced interpretability. This paper introduces a novel framework that challenges this trend by reinterpreting learning from an information-theoretic perspective, viewing it as a search for encoding schemes that capture intrinsic data structures through compact representations. Rather than following the conventional approach of fitting data to complex models, we propose a fundamentally different method that maps data to intervals of initial conditions in a dynamical system. Our GLS (Generalized L\"uroth Series) coding compression classifier employs skew tent maps - a class of chaotic maps - both for encoding data into initial conditions and for subsequent recovery. The effectiveness of this simple framework is noteworthy, with performance closely approaching that of well-established machine learning methods. On the breast cancer dataset, our approach achieves 92.98\% accuracy, comparable to Naive Bayes at 94.74\%. While these results do not exceed state-of-the-art performance, the significance of our contribution lies not in outperforming existing methods but in demonstrating that a fundamentally simpler, more interpretable approach can achieve competitive results. 
\end{abstract}

\begin{IEEEkeywords}
Compression, Generalized L\"uroth Series, Learning, Minimum Description Length, Coding Theory
\end{IEEEkeywords}

\section{Introduction}\label{sec:introduction}
Any data-driven learning fundamentally constitutes an optimization process of encoding the information available in the data. This encoding is motivated by two constraints: maximizing representational fidelity and minimizing encoding length. 
Formally, an optimal encoder $E$ possesses two critical properties: Information Preservation: $E$ faithfully captures the essential statistical characteristics and structural regularities inherent in the data.
Descriptive Efficiency: $E$ enables significant bit-level compression by exploiting fundamental data redundancies and regularities.

The pursuit of such encodings is mathematically formalized through the Minimum Description Length (MDL) principle \cite{rissanen1978modeling} (an information-theoretic reformulation of Occam's razor, \cite{solomonoff1964formal}). MDL attempts to quantify the trade-off between model complexity and representational accuracy by minimizing the total description length—comprising both of the model's complexity and the data encoding.
Conceptually, this reduces to the fact that the most informative model is one that achieves maximal data compression while maintaining predictive capacity.

The MDL principle, although biologically inspired (\cite{motiwala2022efficient, chater2003simplicity}) and known to be theoretically optimal for model selection in machine learning (\citep{hansen2001model}), has not been adopted as the primary motivation for model training and representation learning. Traditionally, model selection often prioritizes alternative criteria for achieving optimal fit, rather than adhering to the principle of minimizing the description length of the model or its encoded representation. Regardless, MDL is the theoretical foundation for common loss functions. The cross-entropy loss used in classification tasks is an MDL optimal encoding of the data given the model (\cite{hinton1993keeping, adriaans2007mdl, marton2005compression}). Specifically, for a probabilistic model $h$, the probability $- \sum_i^N log_2 p(y_{i}|x_{i})$, which corresponds the categorical cross-entropy loss turns out to be the optimal cost associated with transmitting the labels y with this codelength. Similarly, the squared error loss corresponds to MDL coding under Gaussian assumptions (\cite{galland2003mdl, hinton1993keeping}). Despite this, the conventional method of estimating MDL code lengths through variational inference technique originally proposed to minimize description complexity in \cite{hinton1993keeping}—has demonstrated low efficiency (\cite{blier2018description}). Similarly, while techniques such as regularization (\cite{saito1997mdl}) implicitly implement a form of MDL by penalizing overly complex models, fewer approaches target explicit search for simpler models or data compression (\cite{marton2005compression}) during training (\cite{blier2018description, lan2022minimum}). This is particularly challenging because modern learning is synonymous with machine learning and deep learning frameworks which are inherently complex by design, with simplification efforts aiming to sparsify within a vast parameter space.

Thus, algorithmic approaches to data encoding and representation learning currently occupy a small subspace within the landscape of potential learning methodologies that simultaneously prioritize succinctness and informational sufficiency. 
In this paper we challenge the currently pravelant notion of learning from complex models and introduce an extremely simple approach to learning, especially in terms of model complexity and interpretability. We show that by reinterpreting learning from the lens of information theory and Chaos, the complexity of big models can be reduced to search for good initial conditions. Our proposed methodology consists of a model that is described by a very small number of parameters that scale linearly with the number of classes, precisely the parameter count is \(4 \times \text{number of classes}\) (so \(4 \times 3 \) for Iris dataset), with only one hyperparameter for learning. 
This simplification becomes possible through a novel reinterpretation; by treating data generation as a dynamical chaotic system and the datapoints itself as an orbit. Within this framework, the modeling process is reconceptualized as estimating the system's initial conditions necessary to recovering the orbit. Under MDL, the optimal initial condition is one that represents the data with the most compact description length.

\subsection*{Data generation process as a chaotic dynamical system }
As stated before, in this work rather than relying on parametric distributions or complex neural architectures, we reconceptualize the data generating process itself as a chaotic dynamical system. 
This approach leverages fundamental properties of chaotic systems - including topological transitivity, dense periodic orbits, and strange attractors - which naturally align with observed patterns in real-world (\cite{soramaki2007topology, stein2013ecological, palmer1993extended}).
 
This interpretation fundamentally reframes the inference problem: rather than fitting a complex function, we seek to reverse-engineer the interval of initial conditions that gave rise to observed data points through careful backward iteration of the dynamical system. Under the MDL principle, this interval serves as maximally efficient codes for the data, as they capture the generative essence of each observation. 
The contributions of this work are : 
\begin{itemize}
    \item We introduce a novel framework that implements MDL principles through chaotic dynamical systems, reframing data generation and representation learning in terms of initial conditions and their evolution.
    \item Through simple and well-known techniques such as GLS Coding and back-iteration for learning and inference, our method
    \begin{itemize}
        \item Requires significantly less description length for both model specification and data encoding
         \item Provides better interpretability through its compact parameter space.
    \end{itemize}
\item We provide an initial exploration into why chaotic dynamics can effectively capture complex data patterns through simple initial conditions, establishing a connection between dynamical systems theory and efficient data representation.
\end{itemize}

\section{Proposed Method}\label{sec:proposed_method}
In this section, we present the necessary background to understand the proposed method. We begin with an overview of Generalized L\"uroth Series (GLS) coding and explain its Shannon optimality under the assumption of independence. Following this, we introduce the proposed method in detail.
\subsection{GLS Coding}
\begin{figure}[h]
    \centering
    \includegraphics[width=0.35\textwidth]{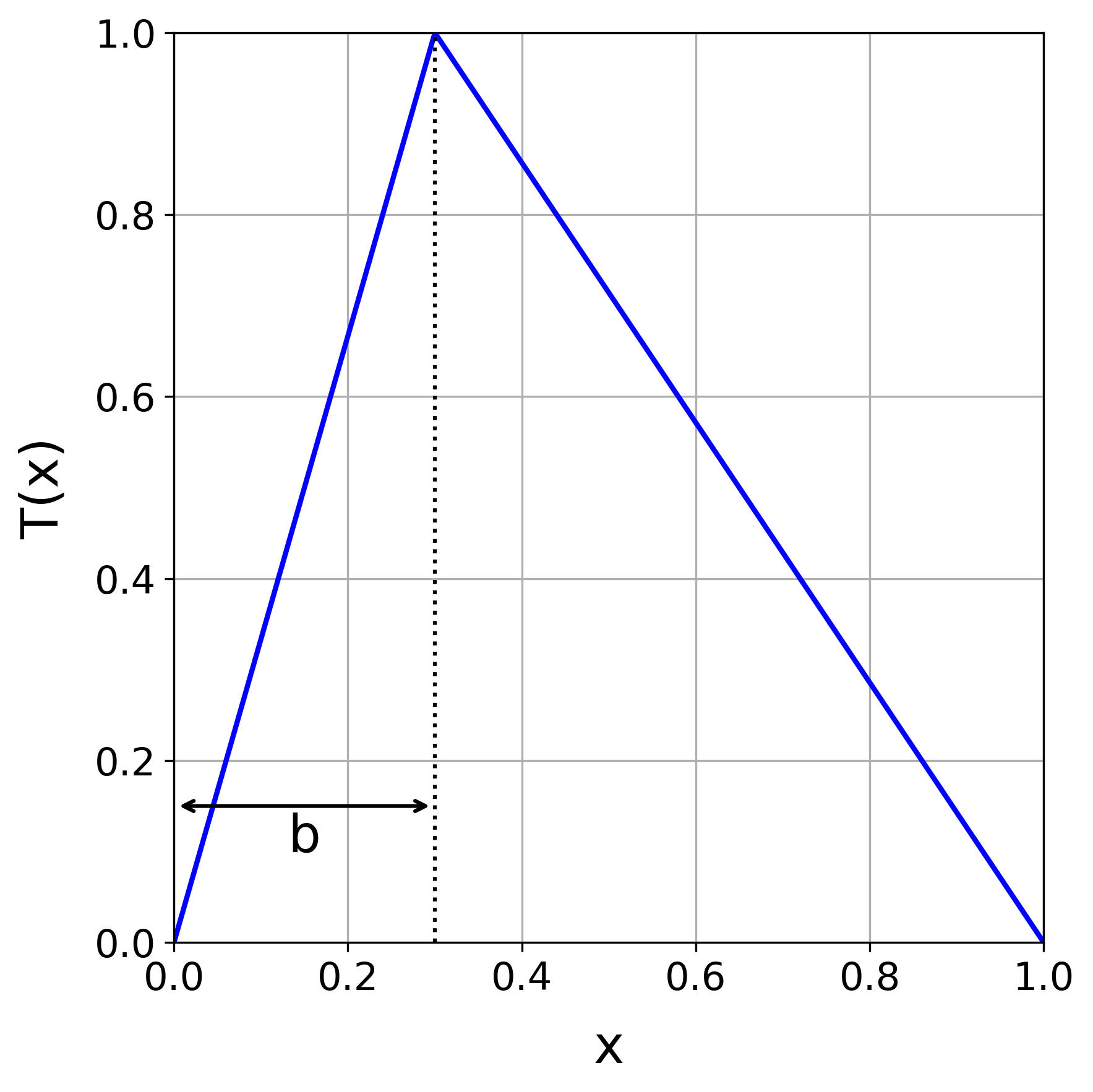}
    \caption{Plot of the skew tent map with \( b=0.3 \).}
    \label{fig:tent_map}
\end{figure}

GLS maps are chaotic, piecewise linear maps. Notable examples include binary maps and skew tent maps. In this work, we employ the skew tent map for GLS coding, defined mathematically as:  

\begin{equation}
T(x) =
\begin{cases} 
\frac{x}{b}, & 0 \leq x < b \\
\frac{(1 - x)}{1-b}, & b \leq x < 1,
\end{cases}
\end{equation}  

where \( b \in (0,1) \) controls the skewness of the tent map. Figure~\ref{fig:tent_map} illustrates the map for \( b = 0.3 \).  

\begin{figure}[h]
    \centering
    \includegraphics[width=0.45\textwidth]{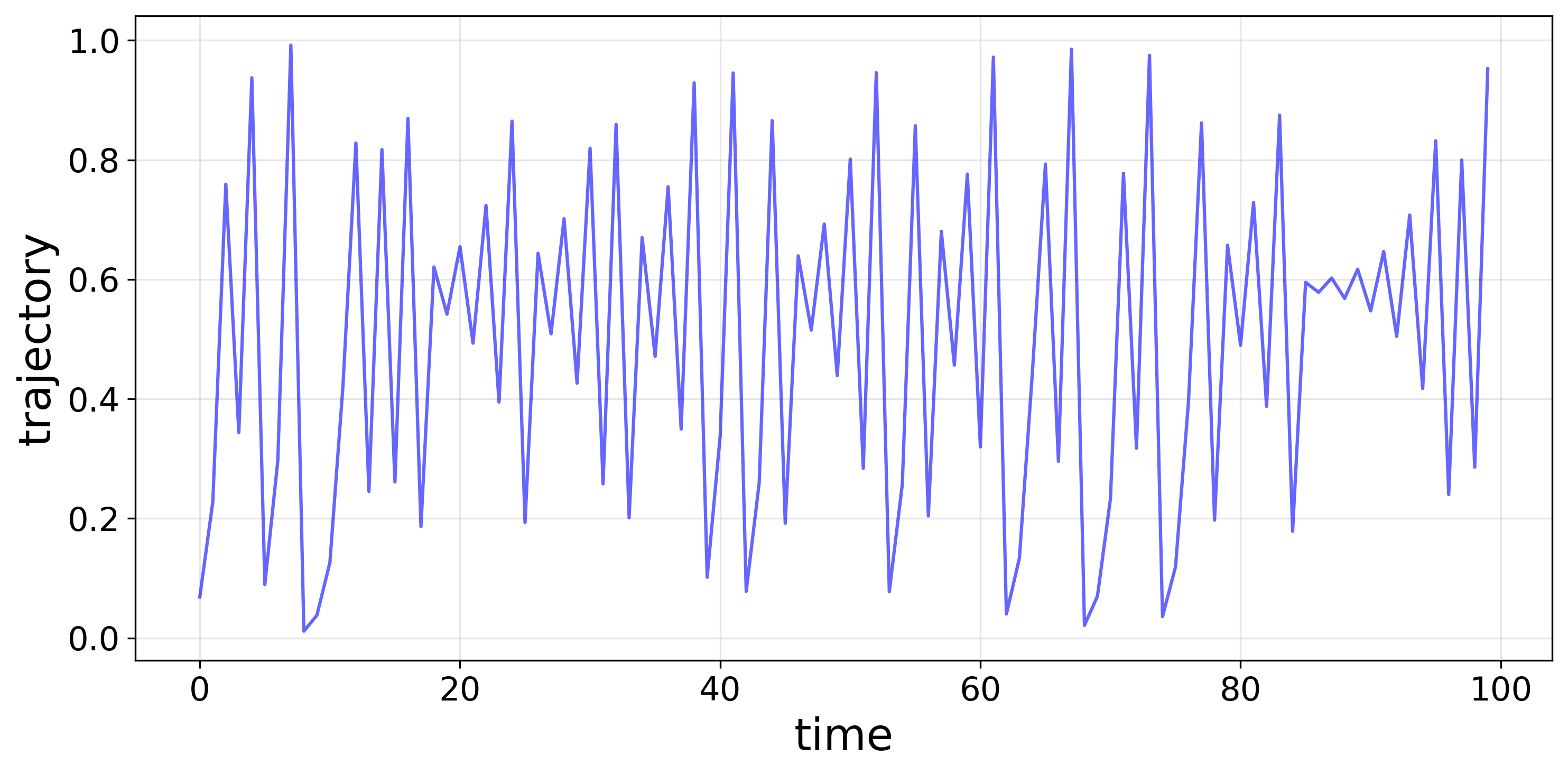}
    \caption{Trajectory of the skew tent map for the first 100 values after removing the initial 500 transients. The parameter \( b \) and initial value \( x_0 \) are 0.34 and 0.26, respectively.}
    \label{fig:tent_map_trajectory}
\end{figure}

The trajectory of the tent map, starting from an initial value \( x_0 \), is defined as:  
\[
x_0 \rightarrow T(x_0) \rightarrow T^2(x_0) \rightarrow \dots \rightarrow T^n(x_0) \rightarrow \dots
\]
where \( T^n(x_0) = T(T^{n-1}(x_0)) \). The trajectory can be denoted as:  
\[
x(t) = [x_0, T(x_0), T^2(x_0), \dots, T^n(x_0), \dots]
\]
This sequence is obtained through \textbf{forward iteration}. Figure~\ref{fig:tent_map_trajectory} depicts the first 100 values of the trajectory after discarding the initial 500 transient iterations.  

The trajectory \( x(t) \) can be mapped to a symbolic sequence \( SS(t) \), defined as:  
\begin{eqnarray}
SS(t_i) =
\begin{cases} 
0, &  x(t_i) < b,  \\
1, & b \leq x(t_i) < 1.\\
\end{cases}
\end{eqnarray}
Given any binary symbolic sequence \( SS(t) \) of length \( N \), we determine the interval containing the initial condition \( x_0 \) using \textbf{backward iteration} on the skew tent map, as outlined in Algorithm~\ref{alg:backiteration}. For backward iteration, the parameter $b$ has to be specified. \begin{algorithm}[!ht]
\caption{Initial Condition Reconstruction via Skew Tent Map Backward Iteration}
\label{alg:backiteration}
\begin{algorithmic}
\Require Symbolic sequence $z = (z_1, z_2, \ldots, z_n)$, parameter $b$
\State $SS \gets (z_n, z_{n-1}, \ldots, z_1)$  \Comment{reverse sequence}
\If{$SS_0 = 0$}
\State $L \gets 0$, $U \gets b$
\Else
\State $L \gets b$, $U \gets 1$
\EndIf
\For{$i \gets 1$ to $n$}
\If{$SS_i = 0$}
\State $L \gets L \cdot b$
\State $U \gets U \cdot b$
\Else
\State $L \gets L \cdot b - L + 1$
\State $U \gets U \cdot b - U + 1$
\EndIf
\If{$L > U$}
    \State swap($L$, $U$)
\EndIf
\EndFor
\State \Return $x_0 \gets \frac{L + U}{2}, \textbf{Interval} [L, U]$
\end{algorithmic}
\end{algorithm}

The skewness parameter \( b \) is chosen as 
\[
b = \frac{n_{\text{zeros}}}{N},
\]

where \( n_{\text{zeros}} \) is the number of zeros in \( SS(t) \).  

Backward iterating on the sequence yields an interval \([L, U]\) such that any \( x_0 \in [L, U] \) reproduces the exact symbolic sequence under forward iteration. A common choice for \( x_0 \) is the midpoint:  
\[
x_0 = \frac{L + U}{2}.
\]
The initial condition \( x_0 \) is then iterated forward under the skew tent map for \( N \) steps, generating a chaotic trajectory \( x(t) \):  
\[
x_0 \rightarrow T(x_0) \rightarrow T^2(x_0) \rightarrow \dots \rightarrow T^N(x_0).
\]
The corresponding symbolic sequence \( SS^*(t) \) is obtained using the thresholding rule:  
\begin{equation}
SS^*(t_i) =
\begin{cases} 
0, &  x(t_i) < b,  \\
1, & b \leq x(t_i) < 1.\\
\end{cases}
\end{equation}
Since \( x_0 \) was chosen from the computed interval, forward iteration for \( N \) steps reconstructs the original sequence exactly, i.e.,  

\[
SS^*(t) = SS(t).
\]

In theory, any value of \( b \in (0,1) \) can be used to reconstruct the symbolic sequence without error. However, as shown in~\cite{nagaraj2009arithmetic}, selecting  

\[
b = \frac{n_{\text{zeros}}}{N}
\]

is Shannon-optimal. This optimality is established by analyzing the number of bits required to encode the interval \([L, U]\), given by $
\lceil -\log_2 (U - L) \rceil$~\cite{nagaraj2009arithmetic}. This can be empirically understood through a simple experiment. Consider that a binary sequence of fixed length (\( N = 20 \)) is generated randomly, where the probability of a zero is set to \( 0.45 \). This ensures that the generated sequences maintain the given probability distribution statistically. The experiment iterates over different values of \( b \) in the range starting from $0.01$ to $0.99$ using 100 linearly spaced values. For each \( b \), $100$ random symbolic sequences are generated. Using the \textbf{backward iteration} method on the skew tent map, the interval \([L, U]\) containing the initial condition is determined. The \textbf{compressed file size} is computed using the formula: $\lceil -\log_2(U - L) \rceil$ as mentioned in~\cite{nagaraj2009arithmetic}. The average compressed size is then calculated across the 100 trials for each \( b \).

\begin{figure}[h]
    \centering
    \includegraphics[width=0.45\textwidth]{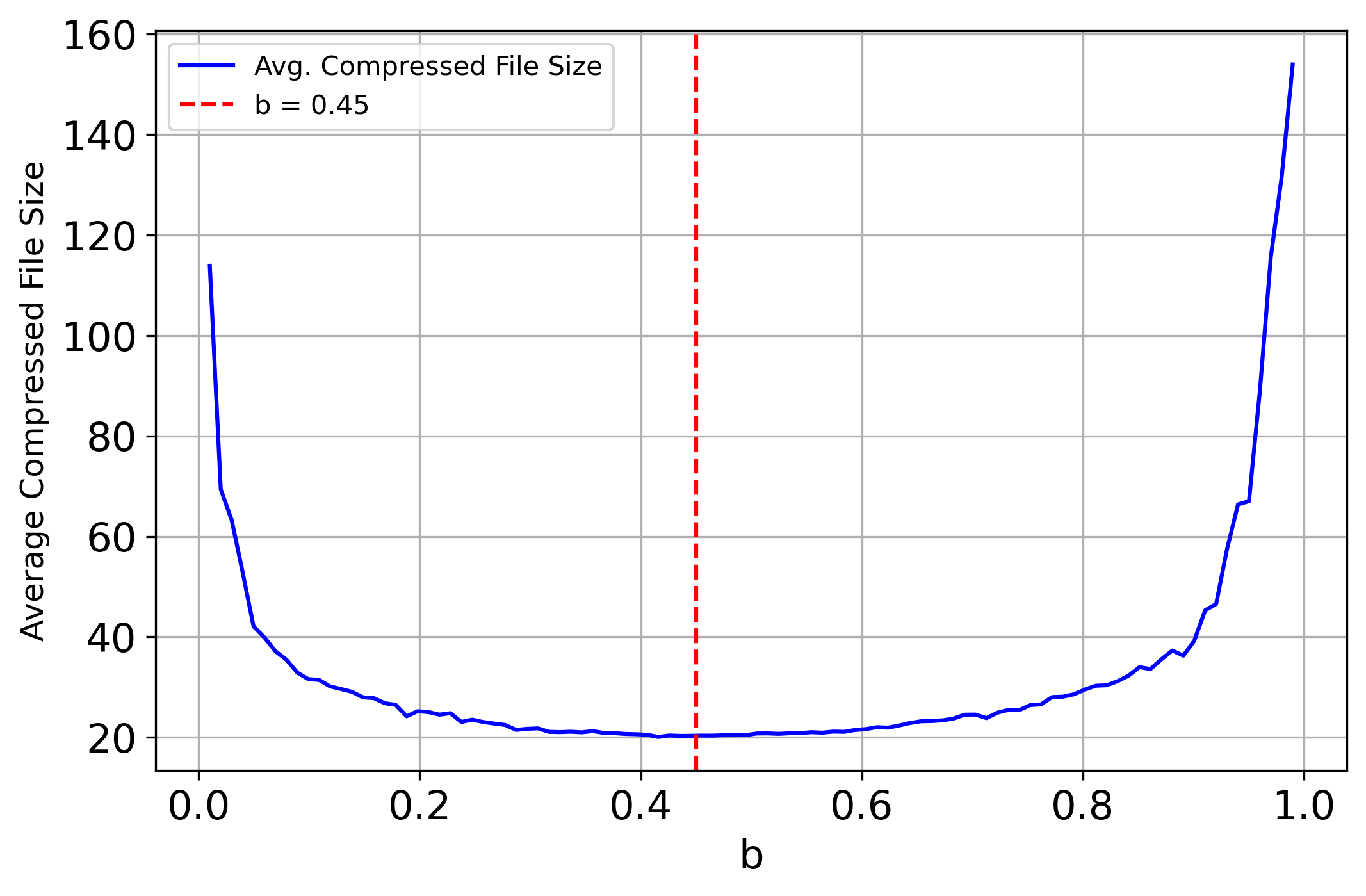}
    \caption{Plot of the average compressed file size as a function of the skew parameter $b$. 
    The red dashed line highlights the optimal value \( b = 0.45 \), corresponding to the empirical probability of zeros in the symbolic sequence.}
    \label{fig:compressed_file_size}
\end{figure}

The graph showing the relationship between average compressed file size vs $b$ is depicted in Figure~\ref{fig:compressed_file_size}. The optimal \( b = 0.45 \) (equal to the probability of zeros in the sequence) is highlighted with a red dashed vertical line. 

The compressed file size is expected to be minimized around this optimal \( b \), as established in prior research on Shannon-optimal coding in~\cite{nagaraj2009arithmetic}. The results illustrate that choosing \( b \) based on symbol probabilities leads to the most efficient encoding, verifying the theoretical foundations of GLS coding.

\section{Proposed Algorithm}
The GLS coding method described in the previous section, which utilizes the skew tent map to encode the interval containing the initial condition, assumes that the symbolic sequence is independently and identically distributed (i.i.d.). However, this assumption is often violated in real-world data, particularly in time series and image data, where strong correlations exist between neighboring values. To account for these dependencies, instead of using the first return map (\(T(x)\) vs. \(x\)), we employ the second return map (\(T(T(x))\) vs. \(x\)). In this approach, the skewness of the map is determined based on the empirical probabilities of non-overlapping consecutive symbol pairs: \(p_{00}\), \(p_{01}\), \(p_{10}\), and \(p_{11}\). These probabilities are estimated from the given symbolic sequence, allowing the GLS coding scheme to better capture the underlying structure of correlated data. 

The equation for the second return skew tent map determined based on the empirical probabilities of non-overlapping consecutive symbol pairs: \(p_{00}\), \(p_{01}\), \(p_{10}\), and \(p_{11}\) obtained from the symbolic sequence is provided below:
\begin{equation*}
T(T(x)) = \begin{cases} 
\dfrac{x}{p_{00}}, & 0 \leq x < p_1 \\[8pt]
\dfrac{p_{00} + p_{01} - x}{p_{01}}, & p_1 \leq x < p_2 \\[8pt]
\dfrac{x - (p_{00} + p_{01})}{p_{11}}, & p_2 \leq x < p_3 \\[8pt]
\dfrac{1 - x}{p_{10}}, & p_3 \leq x < 1
\end{cases}
\end{equation*}
where $p_1 = p_{00}$, $p_2 = p_{00} + p_{01}$, $p_3 = p_{00} + p_{01} + p_{11}$.
\begin{figure}[!tbp]
    \centering
    \includegraphics[width=0.35\textwidth]{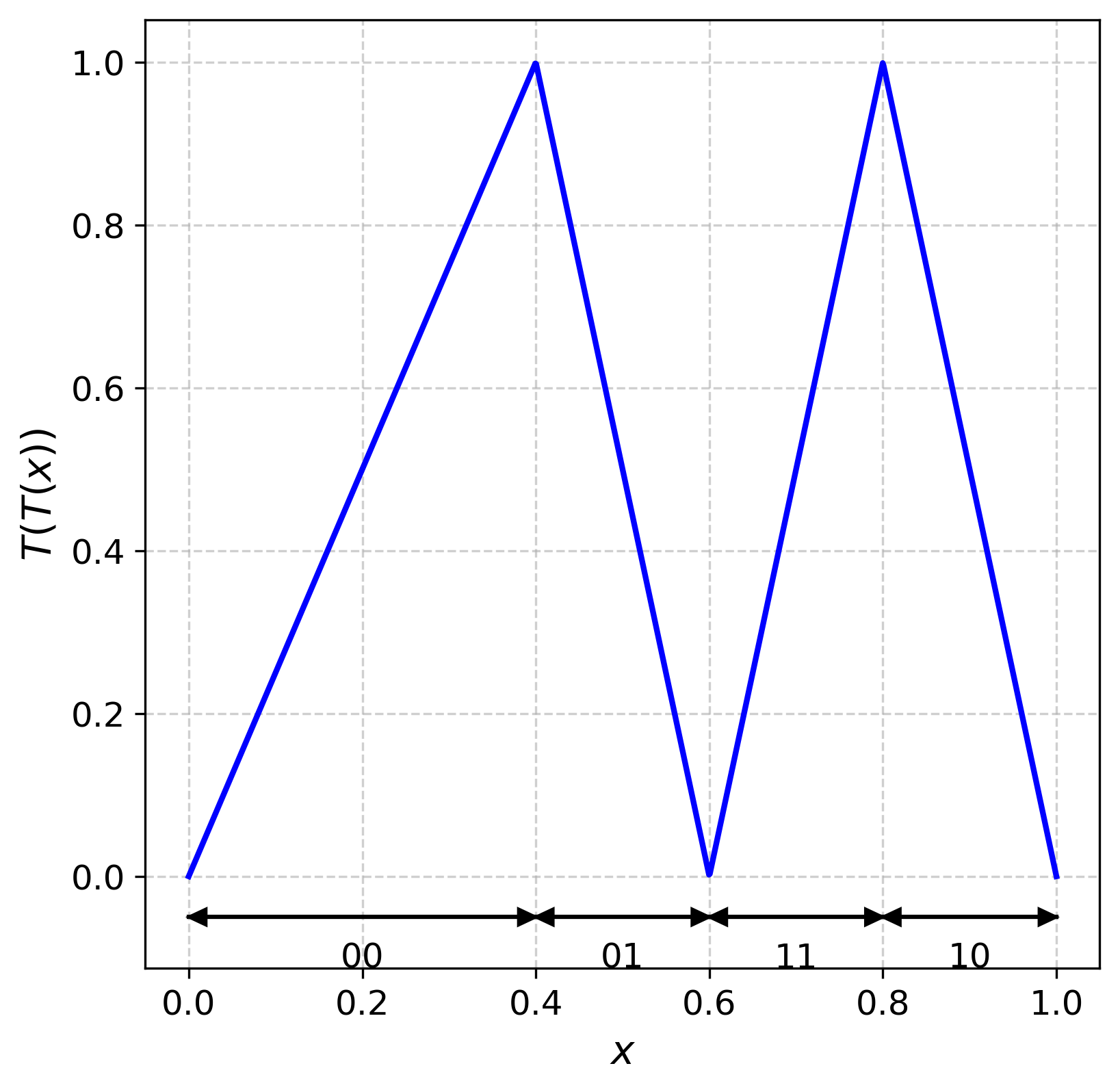}
    \caption{Plot of the second return skew tent map with symbolic sequence regions labeled as \( 00, 01, 11, 10 \). The arrows indicate the corresponding intervals for each sequence. The skewness parameters used are \( p_{00} = \frac{2}{5} \), \( p_{01} = \frac{1}{5} \), \( p_{11} = \frac{1}{5} \), and \( p_{10} = \frac{1}{5} \).}
    \label{fig:second_return_skew_tent_map}
\end{figure}
The second return map assumes that each pair of consecutive, non-overlapping symbolic sequences is independent of the subsequent pairs in the sequences.
For eg. given the symbolic sequence: $SS = [0,0,1,0,0,1,1,1,0,0]$, we extract the non-overlapping consecutive symbol pairs:$
(0,0), (1,0), (0,1), (1,1), (0,0)$. Now we will count their occurrences:
\begin{itemize}
    \item \( (0,0) \) appears 2 times.
    \item \( (0,1) \) appears 1 time.
    \item \( (1,0) \) appears 1 time.
    \item \( (1,1) \) appears 1 time.
\end{itemize}
The total number of pairs is \(5\). The probabilities for the $(i,j)$ is computed as: 
\[ 
p_{ij} = \frac{\text{count}(ij)}{\text{total pairs}}
\]
$p_{00} = \frac{2}{5}$,  $p_{01} = \frac{1}{5}$,
$p_{10} =  \frac{1}{5}$, $p_{11} = \frac{1}{5}$.

Figure~\ref{fig:second_return_skew_tent_map} shows the second return map with the regions mapped as $00, 01, 11, 10$.





\subsection{GLS Coding Compression Classifier Using Second Return Skew Tent Map }

We first describe the back-iteration algorithm for the second-return skew tent map, where the skewness parameters \( p_{00}, p_{01}, p_{11}, p_{10} \) are derived from the given symbolic sequence. Similar to the back-iteration process using the first-return map, this algorithm determines the interval in which the initial condition must lie to generate the desired symbolic sequence.  The algorithm outlined below computes the initial condition, which, when iterated forward and thresholded using the values \( p_{00}, p_{01}, p_{11}, p_{10} \), reconstructs the symbolic sequence in reverse order, which is then reversed again at the end to match the original symbolic sequence.
\subsection{Training: GLS Coding Compression Classifier Using Second Return Skew Tent Map\label{sec:training} }

The training algorithm for the GLS coding compression classifier using the second return skew tent map is as follows. First, the normalized training data\footnote{The normalization is defined as \(X_{\text{norm}} = \frac{X - \min(X)}{\max(X) - \min(X)}\), applied separately to each feature.} is binarized by applying a threshold\footnote{The threshold is a hyperparameter that must be determined through cross-validation.}, converting each feature into a binary sequence. For each class, the algorithm extracts data instances and computes the empirical probabilities \( p_{00}, p_{01}, p_{11}, p_{10} \) by analyzing non-overlapping consecutive symbol pairs in each instance. These probabilities capture the transition frequencies between symbols within class-specific data. To obtain a representative distribution for the class, the algorithm computes the average probabilities across all samples belonging to that class. If any of the average probability of pairs is 0, then we do laplace smoothing. The exact formulation for laplace smoothing applied is provided below:

\[
\hat{p}_{00} = \frac{\sum_{i=1}^{M} p_{i, (0,0)} + \alpha}{M + 4 \cdot \alpha},
\]

\[
\hat{p}_{01}  = \frac{\sum_{i=1}^{M} p_{i, (0,1)} + \alpha}{M + 4 \cdot \alpha},
\]

\[
\hat{p}_{11}  = \frac{\sum_{i=1}^{M} p_{i, (1,1)} + \alpha}{M + 4 \cdot \alpha},
\]

\[
\hat{p}_{10}  = \frac{\sum_{i=1}^{M} p_{i, (1,0)} + \alpha}{M + 4 \cdot \alpha},
\]

where:
\begin{itemize}
  \item \( p_{i, (a,b)} \) is the probability of the pair \((a,b)\) in the \(i^{\text{th}}\) data instance.
  \item \( M \) is the number of data instance of a particular class.
  \item \( \alpha \) is the smoothing parameter added to avoid zero values and ensure stability. In our experiments, we choose $\alpha = 0.001$.
\end{itemize}

The final output is a dictionary containing these averaged probability distributions, which serve as the fundamental statistical parameters for GLS-based compression and classification. The complete training procedure is outlined in Algorithm~\ref{alg:GLS_Coding}. Another thing to note is that this compressed dictionary, $size = \text{No. of classes} \times 4$ is the total size of our model, significantly smaller than most machine learning models.
 
\begin{algorithm}[!h]
\caption{Back Iteration Algorithm for Second-Return Skew Tent Map}
\label{alg:back_iteration}
\begin{algorithmic}[1]
\Require \( p_{00}, p_{01}, p_{11}, p_{10} \) (Second Return Skew Tent Map Parameters), \( \mathbf{S}=[S_1, S_2, \ldots, S_N] \) (Symbolic Sequence)
\State Initialize interval \( [L, U] \) as an empty array
\State \textbf{Extract} consecutive non-overlapping symbol pairs from \( \mathbf{S} \)

\State \textbf{Set initial interval} based on the first symbol pair:
\If{\( (S_1, S_2) = (0,0) \)}
    \State \( L \gets 0 \), \( U \gets p_{00} \)
\ElsIf{\( (S_1, S_2) = (0,1) \)}
    \State \( L \gets p_{00} \), \( U \gets p_{00} + p_{01} \)
\ElsIf{\( (S_1, S_2) = (1,1) \)}
    \State \( L \gets p_{00} + p_{01} \), \( U \gets p_{00} + p_{01} + p_{11} \)
\ElsIf{\( (S_1, S_2) = (1,0) \)}
    \State \( L \gets 1 - (p_{00} + p_{01} + p_{11}) \), \( U \gets 1 \)
\EndIf

\For{each subsequent pair \( (S_i, S_{i+1}) \) in \( \mathbf{S}, \) where $i$ starts from $3$}
    \If{\( (S_i, S_{i+1}) = (0,0) \)}
        \State \( L \gets L \cdot p_{00} \), \( U \gets U \cdot p_{00} \)
    \ElsIf{\( (S_i, S_{i+1}) = (0,1) \)}
        \State \( L \gets p_{00} + p_{01} - p_{01} \cdot L \)
        \State \( U \gets p_{00} + p_{01} - p_{01} \cdot U \)
    \ElsIf{\( (S_i, S_{i+1}) = (1,1) \)}
        \State \( L \gets p_{11} \cdot L + p_{00} + p_{01} \)
        \State \( U \gets p_{11} \cdot U + p_{00} + p_{01} \)
    \ElsIf{\( (S_i, S_{i+1}) = (1,0) \)}
        \State \( L \gets 1 - p_{10} \cdot L \)
        \State \( U \gets 1 - p_{10} \cdot U \)
    \EndIf
    \If{\( L > U \)}
        \State Swap \( L \) and \( U \)
    \EndIf
\EndFor

\State \Return \( x_0 = \frac{L + U}{2} \), Interval \( [L, U] \)
\end{algorithmic}
\end{algorithm}

\subsection{TESTING: GLS CODING COMPRESSION CLASSIFIER USING SECOND RETURN SKEW TENT MAP}
During the testing phase of the GLS coding compression classifier, each test instance is first binarized using the same threshold applied during training. Given \( n \) classes, there exist \( n \) sets of average empirical probabilities \( \bar{p}_{00}, \bar{p}_{01}, \bar{p}_{11}, \bar{p}_{10} \), where each set corresponds to a specific class and is precomputed from the training data. For a given test instance, the classifier performs back-iteration separately for each class using its respective probability set, thereby computing the interval \([L, U]\). The compressed file size for each class is then determined as  
\[
\lceil- \log_2 (U - L) \rceil.
\]
The class label is assigned based on the smallest compressed file size, i.e., the predicted class \( c^* \) is given by  

\[
c^* = \arg\min_{c} \lceil -\log_2 (U_c - L_c) \rceil,
\]

where \( U_c \) and \( L_c \) denote the interval bounds computed for class \( c \). This approach ensures that the test instance is classified into the class that achieves the most efficient compression. The algorithm for testing is shown in Algorithm~\ref{alg:GLS_Coding_Predict}.

\begin{algorithm}[!h]
\caption{GLS Coding Training using Second Return Map with Laplace Smoothing}
\label{alg:GLS_Coding}
\begin{algorithmic}[1]
    \Require Normalized training data $X_{train}^{norm}$, Training labels $y_{train}$, Threshold $\tau$, Laplace smoothing parameter $\alpha$
    \Ensure Average class probabilities $\mathcal{P}_{avg}$
    \State Convert training data to binary: $X_{train}^{bin} \gets (X_{train}^{norm} \geq \tau)$
    \State Identify number of classes: $n_{classes} \gets |\text{unique}(y_{train})|$
    \State Initialize $\mathcal{P}_{avg} \gets \emptyset$
    \For{each class $c$ in $\{0,1,\dots,n_{classes}-1\}$}
        \State Select class samples: $X_c \gets X_{train}^{bin}[y_{train} = c]$
        \State Initialize probability list $\mathcal{P}_c \gets \emptyset$
        \For{each sample $x \in X_c$}
            \State $\text{Compute pair probabilities}^*$ $p_{00}, p_{01}, p_{11}, p_{10} \gets \text{Compute Pair Probabilities}(x)$
            \State Append $\{(0,0): p_{00}, (0,1): p_{01}, (1,1): p_{11}, (1,0): p_{10}\}$ to $\mathcal{P}_c$
        \EndFor
        \State Compute average probabilities:
        \State $\bar{p}_{00} \gets \frac{1}{|X_c|} \sum_{p \in \mathcal{P}_c} p_{00}$
        \State $\bar{p}_{01} \gets \frac{1}{|X_c|} \sum_{p \in \mathcal{P}_c} p_{01}$
        \State $\bar{p}_{11} \gets \frac{1}{|X_c|} \sum_{p \in \mathcal{P}_c} p_{11}$
        \State $\bar{p}_{10} \gets \frac{1}{|X_c|} \sum_{p \in \mathcal{P}_c} p_{10}$

        \If{$\bar{p}_{00} = 0 \lor \bar{p}_{01} = 0 \lor \bar{p}_{11} = 0 \lor \bar{p}_{10} = 0$}
            \State \textbf{Laplace smoothing:}
            \State $\hat{p}_{00} \gets \frac{\sum_{p \in \mathcal{P}_c} p_{00} + \alpha}{|X_c| + 4 \alpha}$
            \State $\hat{p}_{01} \gets \frac{\sum_{p \in \mathcal{P}_c} p_{01} + \alpha}{|X_c| + 4 \alpha}$
            \State $\hat{p}_{11} \gets \frac{\sum_{p \in \mathcal{P}_c} p_{11} + \alpha}{|X_c| + 4 \alpha}$
            \State $\hat{p}_{10} \gets \frac{\sum_{p \in \mathcal{P}_c} p_{10} + \alpha}{|X_c| + 4 \alpha}$
        \Else
            \State $\hat{p}_{00} \gets \bar{p}_{00}$
            \State $\hat{p}_{01} \gets \bar{p}_{01}$
            \State $\hat{p}_{11} \gets \bar{p}_{11}$
            \State $\hat{p}_{10} \gets \bar{p}_{10}$
        \EndIf

        \State Store class probability: $\mathcal{P}_{avg}[c] \gets \{(0,0): \hat{p}_{00}, (0,1): \hat{p}_{01}, (1,1): \hat{p}_{11}, (1,0): \hat{p}_{10}\}$
    \EndFor
    \State \Return $\mathcal{P}_{avg}$
\end{algorithmic}
\vspace{0.5em}
{\small \textbf{$^*$} Compute Pair Probabilities includes the calculation of the joint probability for pairs of class outcomes, such as $(0,0), (0,1), (1,1), (1,0)$ for each sample.}
\end{algorithm}

\begin{algorithm}[!h]
\caption{GLS Coding Compression Classifier - Prediction Phase}
\label{alg:GLS_Coding_Predict}
\begin{algorithmic}[1]
\Require $X_{\text{test\_norm}}$: Normalized test data, $n_{\text{classes}}$: Number of classes, 
$\mathcal{P}_{avg}$: Dictionary of average class probabilities, $\tau$: Threshold for binarization
\Ensure Predicted class labels for each test instance

\State Binarize $X_{\text{test\_norm}}$ using threshold $\tau$: 
\[
X_{\text{test\_bin}} = (X_{\text{test\_norm}} \geq \tau)
\]

\State Initialize matrix $S_{\text{test}} \in \mathbb{R}^{m \times n_{\text{classes}}}$ to store compressed sizes 

\For{each test instance $x_i \in X_{\text{test\_bin}}$} 
    \For{each class label $c \in \{0, \dots, n_{\text{classes}} - 1\}$}
        \State Retrieve average probabilities $\hat{p}_{00}, \hat{p}_{01}, \bar{p}_{11}, \hat{p}_{10}$ from $\mathcal{P}_{avg}[c]$
        \State Perform back iteration on second return skew tent map to compute interval:
        \[
        [L_c, U_c] = \text{back\_iterate}(\hat{p}_{00}, \hat{p}_{01}, \hat{p}_{11}, \hat{p}_{10}, x_i)
        \]
        \State Compute compressed file size:
        \[
        S_{\text{test}}[i, c] = \lceil -\log_2 (U_c - L_c) \rceil
        \]
    \EndFor
\EndFor

\State Assign each test instance the class with the smallest compressed file size:
\[
\hat{y}_i = \arg\min_{c} S_{\text{test}}[i, c]
\]

\State \Return Predicted class labels $\hat{y}$
\end{algorithmic}
\end{algorithm}
\section{Experiments and Results}
We conducted experiments on six datasets, the \emph{Iris} (\cite{fisher1936iris}), \emph{Breast Cancer Wisconsin} (\cite{breast_cancer_wisconsin_(diagnostic)_17}), \emph{Wine} (\cite{wine_109}), and \emph{Bank Note Authentication} (\cite{banknote_authentication_70}), \emph{Ionosphere} \cite{ionosphere_49}, \emph{Seeds} (\cite{seeds_236}) and \emph{Haberman's Survival dataset} (\cite{haberman1973analysis}). Each dataset was split into 80\% for training and 20\% for testing, with a fixed random seed of 42 to ensure reproducibility. The number of training and test instances per class for each dataset is summarized in Table~\ref{tab:train_test_instances}.
\textbf{Iris :}
Using five-fold cross-validation with a fixed random state of 90, we optimized the threshold for macro F1-score by varying it from 0.01 to 1.00 in steps of 0.01. The best threshold (0.59) yielded an average macro F1-score of 0.73. We have applied laplace smoothing to the model for Iris data.
\textbf{Breast Cancer Wisconsin :}
Following the same procedure, the optimal threshold (0.32) achieved an average macro F1-score of 0.92.   
\textbf{Wine : }
With an optimal threshold of 0.25, cross-validation resulted in an average macro F1-score of 0.812. 
\textbf{Bank Note Authentication :}
With an optimal threshold of 0.62, cross-validation resulted in an average macro F1-score of 0.77. 
\textbf{Ionosphere :}
With an optimal threshold of 0.01, cross-validation resulted in an average macro F1-score of 0.82. 
\textbf{Seeds : }
With an optimal threshold of 0.51, cross-validation resulted in an average macro F1-score of 0.85.
\textbf{Haberman's Survival :}
With an optimal threshold of 0.61, cross-validation resulted in an average macro F1-score of 0.52.
The performance results for all datasets are summarized Table~\ref{tab:classification_results}. 

\begin{table*}
    \centering
    \caption{Classification Performance on Different Datasets}
    \label{tab:classification_results}
    \begin{tabular}{lccccc}
        \toprule
        Dataset & Accuracy & Macro Precision & Macro Recall & Macro F1-score & Best Threshold \\
        \midrule
        Iris\tablefootnote{Laplace Smoothing is applied.} & 0.8667 & 0.8953 & 0.8519 & 0.8468 & 0.5900 \\
         Breast Cancer Wisconsin & 0.9298 & 0.9292 & 0.9207 & 0.9246 & 0.3200 \\
       Wine\tablefootnote{Laplace Smoothing is applied.} & 0.7500 & 0.7357 & 0.7500 & 0.7387 & 0.2500 \\
       Bank Note Authentication & 0.7927 & 0.7923 & 0.7940 & 0.7923 & 0.6200 \\
       Ionosphere & 0.8028 & 0.8772 & 0.7500 & 0.7633 & 0.0100 \\
       Seeds\tablefootnote{Laplace Smoothing is applied.} & 0.7857 & 0.7871 & 0.7872 & 0.7789 & 0.5100 \\
       Haberman's Survival & 0.6774 & 0.6129 & 0.6250 & 0.6206 & 0.6100\\
        \bottomrule
    \end{tabular}
\end{table*}

\begin{table}[!h]
    \centering
    \caption{Class Distribution in Train and Test Sets}
    \label{tab:train_test_instances}
    \begin{tabular}{l|c|c}
        \hline
        \textbf{Dataset} & \textbf{Train Instances} & \textbf{Test Instances} \\
        \hline
        Iris & (40,41,39) & (10,9,11) \\
        Breast Cancer & (169,286) & (43,71) \\
        Wine & (45,57,40) & (14,14,8) \\
        Bank Note & (614,483) & (148,127) \\
        Ionosphere & (98,182) & (28,43) \\
        Seeds & (59,56,53) & (11,14,17) \\
        Haberman & (181, 63) & (44, 18)\\
        \hline
    \end{tabular}

\end{table}
\subsection{ Connection with MDL optimality}

In this section, we make connections between our approach and principles of MDL for optimal encodings. Let \( \mathcal{H} \) be the hypothesis space consisting of all second return maps parameterized by transition probabilities \( (p_{00}, p_{01}, p_{11}, p_{10}) \). The optimal hypothesis \( h^* \in \mathcal{H} \) is one that best explains the data while maintaining simplicity. Using the MDL principle, we aim to minimize the total description length:
\[
h^* = \underset{h \in \mathcal{H}}{\arg\min} \, [L(D|h) + L(h)].
\]
where  \( L(D|h) \) is the description length of the data given hypothesis \( h \) and \( L(h) \) is the description length of the hypothesis itself.
This corresponds to Maximum a Posteriori (MAP) estimation:
\[
h^* = \underset{h \in \mathcal{H}}{\arg\max} \, P(h \mid \mathcal{D}).
\]
Using Bayes' theorem:
\[
h^* = \underset{h \in \mathcal{H}}{\arg\max} \, P(\mathcal{D} \mid h) P(h).
\]
Taking the negative log:
\[
h^* = \underset{h \in \mathcal{H}}{\arg\min} \, -\log_2 P(\mathcal{D} \mid h) - \log_2 P(h).
\]
Now, using the second return map formulation:
\[
P(\mathcal{D} \mid h) \propto U - L.
\]
Thus, the total description length is:
\[
L_{\text{total}} = -\log_2 (U - L ) - \log_2 (P(h)).
\]
where \( h \) in this context is the second return skew tent map. We can assume a uniform prior to $P(h)$.

This formulation ensures that the choice of the best hypothesis aligns with both the data likelihood and the structure imposed by the second return map.

\subsection{Limitations and Future Work}
While our preliminary work demonstrates promising results, several limitations and opportunities for improvement exist. A fundamental limitation of our approach lies in its dependence on computational precision when performing back-iteration to determine intervals of initial conditions. For high-dimensional data (represented by long symbolic sequences), the limited precision available in standard computing environments may affect our ability to resolve the correct intervals. However, this limitation could potentially be addressed through renormalization techniques (\cite{sayood2017introduction}), which we plan to explore in future work. Further, the current implementation restricts the hypothesis class to skew tent maps. Future research could explore other classes of chaotic maps for potential performance improvements, as well as alternative coding algorithms. Finally, another key area is for future work is to enhance model performance while maintaining strict bounds on complexity.

\section{Conclusion}
This work emphasizes on the need for alternative frameworks for learning, ones that put more emphasis on compressibility, interpretability and computational efficiency. To this end, this paper is fundamental in proposing  a method that views learning not as a search for the `best fit' model in the limit of data samples problem, but as a search for minimal yet efficient encoding problem. We leverage fundamentals from information theory for compression and Chaos for non-linear dynamics as those observed in real world problems and data and bring them together to into a novel GLS coding compression classifier. Our GLS coding compression classifier demonstrates that sophisticated pattern recognition can emerge from remarkably simple mechanisms. By mapping data to intervals of initial conditions, we achieve classification performance comparable to established machine learning algorithms - achieving 92.98\% accuracy on breast cancer detection compared to 94.74\% with Naive Bayes, while maintaining significantly lower computational complexity and high interpretability.
The framework approaches traditional machine learning performance while prioritizing simplicity and transparency suggests promising directions for sustainable AI development. Rather than pursuing ever-increasing model complexity and computational demands, we show how fundamental principles from information theory and dynamical systems can yield efficient, interpretable solutions to real-world classification problems.

\section*{Code Availability}
The python code used in this research are available in the following GitHub repository: 
 \href{https://github.com/i-to-the-power-i/gls-coding-classifier}{https://github.com/i-to-the-power-i/gls-coding-classifier}

\section*{Acknowledgements}
Harikrishnan N. B. gratefully acknowledges the financial support provided by the New Faculty Seed Grant (NFSG/GOA/2024/G0906) from BITS Pilani, K. K. Birla Goa Campus. Anuja Vats acknowledges the support from Research Council of Norway, under CapsNetwork Project (Project No:322600) for this work.




\ifarxiv

\end{document}